\def\year{2022}\relax
\documentclass[letterpaper]{article} % DO NOT CHANGE THIS

\usepackage{microtype}
\usepackage{subfigure}
\usepackage{booktabs} % for professional tables
\usepackage{cprotect}
\usepackage{amsmath, amssymb}
\usepackage{listings}

\usepackage{aaai22}  % DO NOT CHANGE THIS
\usepackage{times}  % DO NOT CHANGE THIS
\usepackage{helvet}  % DO NOT CHANGE THIS
\usepackage{courier}  % DO NOT CHANGE THIS
\usepackage[hyphens]{url}  % DO NOT CHANGE THIS
\usepackage{graphicx} % DO NOT CHANGE THIS
\usepackage{natbib}  % DO NOT CHANGE THIS AND DO NOT ADD ANY OPTIONS TO IT
\usepackage{caption} % DO NOT CHANGE THIS AND DO NOT ADD ANY OPTIONS TO IT
\DeclareCaptionStyle{ruled}{labelfont=normalfont,labelsep=colon,strut=off} % DO NOT CHANGE THIS
\usepackage{algorithm}
\usepackage{algorithmic}

\usepackage{newfloat}
\usepackage{listings}
\floatstyle{ruled}
\newfloat{listing}{tb}{lst}{}
\floatname{listing}{Listing}
\title{Training and Evaluating a Jupyter Notebook Data Science Assistant}

\author{
    Shubham Chandel, Colin B. Clement, Guillermo Serrato, and Neel Sundaresan
}
\affiliations{
One Microsoft Way
Redmond, WA 98052
}

\begin{document}

\maketitle

\begin{abstract}
We study the feasibility of a Data Science assistant powered by a sequence-to-sequence transformer by training a new model JuPyT5 on all publicly available Jupyter Notebook GitHub repositories and developing a new metric: Data Science Problems (DSP). DSP is a collection of 1119 problems curated from 306 pedagogical notebooks with 92 dataset dependencies, natural language and Markdown problem descriptions, and assert-based unit tests. These notebooks were designed to test university students' mastery of various Python implementations of Math and Data Science, and we now leverage them to study the ability of JuPyT5 to understand and pass the tests. We analyze the content of DSP, validate its quality, and we find that given 100 sampling attempts JuPyT5 is able to solve 77.5\% of the DSP problems. We further present various ablation and statistical analyses and compare DSP to other recent natural language to code benchmarks.

%We introduce JuPyT5, a text-to-text transformer model, trained on all the publicly available Jupyter Notebooks on Github. The inherent interdependence of code cells in notebooks, provide a challenging environment for code language models to test their long context generation abilities. To this end, we evaluate JuPyT5 on a new benchmark, Data Science Problems (DSP), a collection of 1119 problems, embedded within 306 notebooks. The notebooks in DSP were designed to test a student's understanding of data science and machine learning in the context of assignments presented in universities. In our case, JuPyT5 will act as a student and will be tasked to generate solutions to Data Science problems in DSP to pass an assignment. JuPyT5 is able generate 94.6\% syntactically correct code and is able to solve 37.9\% of all the problems, given 100 chances, indicating the complexity of DSP. Further, we demonstrate pre-training on a markdown rich subset of the data improves performance on DSP to 43.8\%. Finally, we explore the learning from examples behavior and show that notebooks gives a natural environment to learn from examples. 
\end{abstract}

One focus of machine learning research is to build intelligent assistants which can fill-in or predict information based on a context provided by a user. These agents can be as simple as next-word or phrase prediction like in GMail Smart Compose~\cite{chen2019gmail} or as complex as conversational agents based on language models like GPT-3~\cite{brown2020language}. Much work has been done to use these agents to solve natural language tasks, and more recently, to solve software engineering tasks. For example \citet{gptc} offers line-completion to a user and the Codex model~\cite{chen2021evaluating} offers complete methods and classes. While these works evaluate general-purpose coding, there is an opportunity to focus on an agent which offers suggestions in a pedagogical environment: can we develop an agent which can offer students suggestions to solve problems in data science via Jupyter notebooks?

Large language models~\cite{radford2018improving} and transformers~\cite{lewis2019bart} have unlocked consistent improvements~\cite{kaplan2020scaling,brown2020language} in natural language processing and more recently in code synthesis from natural language and examples~\cite{chen2021evaluating,austin2021program,clement2020pymt5}, code completion~\cite{gptc,svyatkovskiy2019pythia,raychev2014code, bruch2009learning}, code search~\cite{husain2019codesearchnet,feng2020codebert}, bug fixing~\cite{deepdebug_java} and detection~\cite{zhai2020cpc}, unit test generation~\cite{tufano2020generating}, and many other applications.

Extending the evaluation of these transformers beyond traditional NLP metrics like BLEU/ROUGE scores, \citet{chen2021evaluating} and \citet{austin2021program} introduced HumanEval and Mostly Basic Programming Problems (MBPP), respectively, which are sets of natural language descriptions of programs along with unit tests and ground truth Python implementations. By compiling and executing generated hypotheses for these Python programs, these works established generally that larger models solve more problems and that drawing more samples (giving the models more attempts) can solve more problems. Further, \citet{austin2021program} showed that human-in-the-loop feedback with model hypotheses could help the model overcome incorrect solutions. Both works observed generally that models struggled to compose descriptions of multiple chained operations.

\begin{figure}[htbp]
    \centering
    \includegraphics[width=\columnwidth]{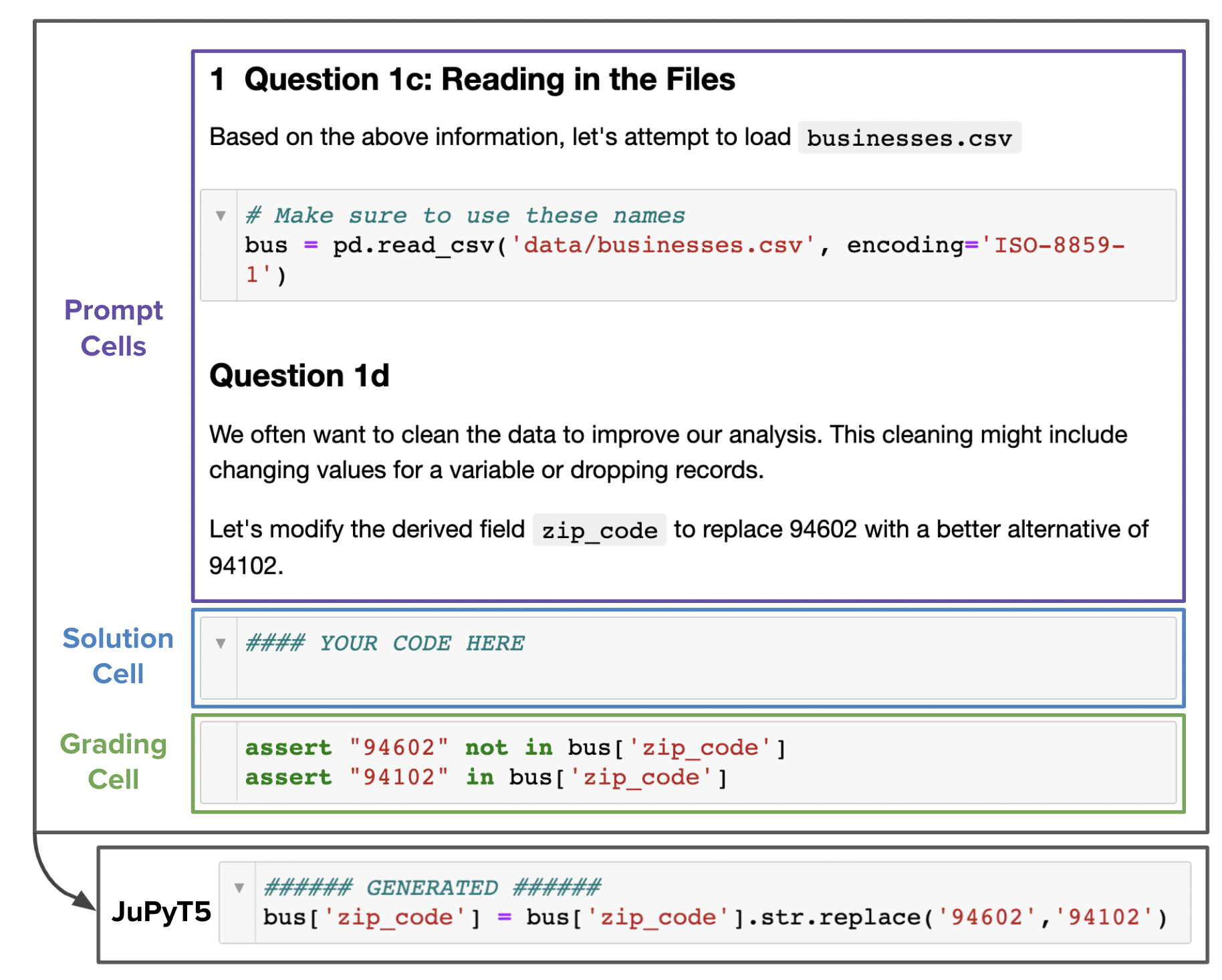}
    \caption{An example (top) from Data Science Problems which loads a file and suggests a data-cleaning modification of the Pandas \texttt{dataframe}, where a solution is to be inserted between the prompt cell and assert-laden grading cell. Given the prompt and context, our model JuPyT5 correctly interprets and implements this code (bottom).}
    \label{fig:data-example}
\end{figure}

Inspired by these evaluations of natural language modeling of source code, this paper introduces an executable Jupyter notebook metric which tests a models ability to solve data science and college-level Computer Science problems. Jupyter notebooks are hybrid code and documentation environments, organized by cells, containing rich Markdown cells, presentation cells, code cells, and output cells. Jupyter notebooks are used widely in both education and business, promoting easily shareable self-documented code in one user experience. As such, Jupyter notebooks are an important environment to test the efficacy of code generation and program language understanding. This paper offers the following contributions:
\begin{enumerate}
    \item We introduce a new evaluation called Data Science Problems (DSP)\footnote{\texttt{github.com/microsoft/DataScienceProblems}}, a curated set of pedagogical Python notebooks and data contexts containing rich Markdown descriptions of problems, solutions, and unit tests, which uses the teaching tool \texttt{nbgrader}\footnote{\texttt{https://nbgrader.readthedocs.io/en/stable/}} to automatically evaluate model hypotheses. DSP problem descriptions also contains natural language with unit tests, featuring \LaTeX~ and math, data-dependencies, and implicit dependencies between the problems in a single notebook.
    
    \item We introduce a new 'code-infilling' pre-training objective, similar to span-mask pre-training and the method-feature-filling objective of PyMT5~\cite{clement2020pymt5}, wherein each cell in each notebook is considered a target in one example, and the source is the neighboring cells with a control code indicating which cell type to produce and where it should be inserted. The resulting model trained by this objective is called JuPyT5 -- Jupyter Python Text-to-text Transfer Transformer.
    
    \item While we do not follow the trend of exploring model size, we focus on evaluating a model size with more modest deployment cost by training and evaluating one model size of 350M parameters (300M non-embedding parameters). We evaluate this model size trained on the cell-infilling objective for our new DSP metric, showing it can solve 78\% of the DSP tasks given 100 sampled attempts. We find similarly that model performance improves with larger number of samples, and more context cells improves the model performance. Surprisingly, training the model with the ability to look ahead a single cell doubles the performance on DSP compared to a 3-cell look-back baseline. Showing the model unit tests also improves performance, and interestingly, the model learns to adapt the solutions to previous problems on a subsequent problem. 
    
    \item We also evaluate JuPyT5 on HumanEval and MBPP. JuPyT5 can beat a much larger 68B parameter model at MBPP with a modest 300M parameters, but was also explicitly trained only on code domain sources. While large models are very impressive general-learners from diverse data sources, it is still much more economical to focus a model training on a task domain. JuPyT5 was outperformed by a similar-sized Codex model on HumanEval, but we were able to partially close the gap by adapting HumanEval docstrings to look more like Markdown. We conclude naturally that smaller models are more formatting-sensitive.
    
    \item We evaluate our model trained on all naturally occurring types of Jupyter notebooks on GitHub, and also on a model trained on a subset of the notebooks containing large amounts of Markdown cells, showing improvement on DSP by balancing the training data to contain similar amounts of code and Markdown.

\end{enumerate}

Data Science Problems are inspired by several existing code execution evaluations in the literature. The first is the APPS dataset~\cite{hendrycks2021measuring}, a collection of 10,000 problems from code competitions. We did not evaluate on this evaluation as we were interested in more task-grounded domains. The second is HumanEval, introduced with the code-fine-tuned GPT-3 model Codex~\cite{chen2021evaluating}, which features Python signatures containing \texttt{doctest} unit tests and natural language docstring descriptions of problems to be solve. The third is Mostly Basic Programming Problems (MBPP), which are  natural language descriptions of problems along with assert-based unit tests. HumanEval and MBPP are more similar to our new DSP metric, but differ a lot in the task domain. DSP is task-grounded as 35\% of its problems explicitly depend on a data context. Further, DSP is an educational domain, so can evaluate the ability of a model to potentially assist in student coaching or continued data science education in business environments. DSP is also larger than HumanEval and MBPP, with over 1000 problem-test pairs, and can test the ability of a model to understand sequences of tasks in context with one another.

\begin{figure}[htbp]
\centering
\includegraphics[width=\columnwidth]{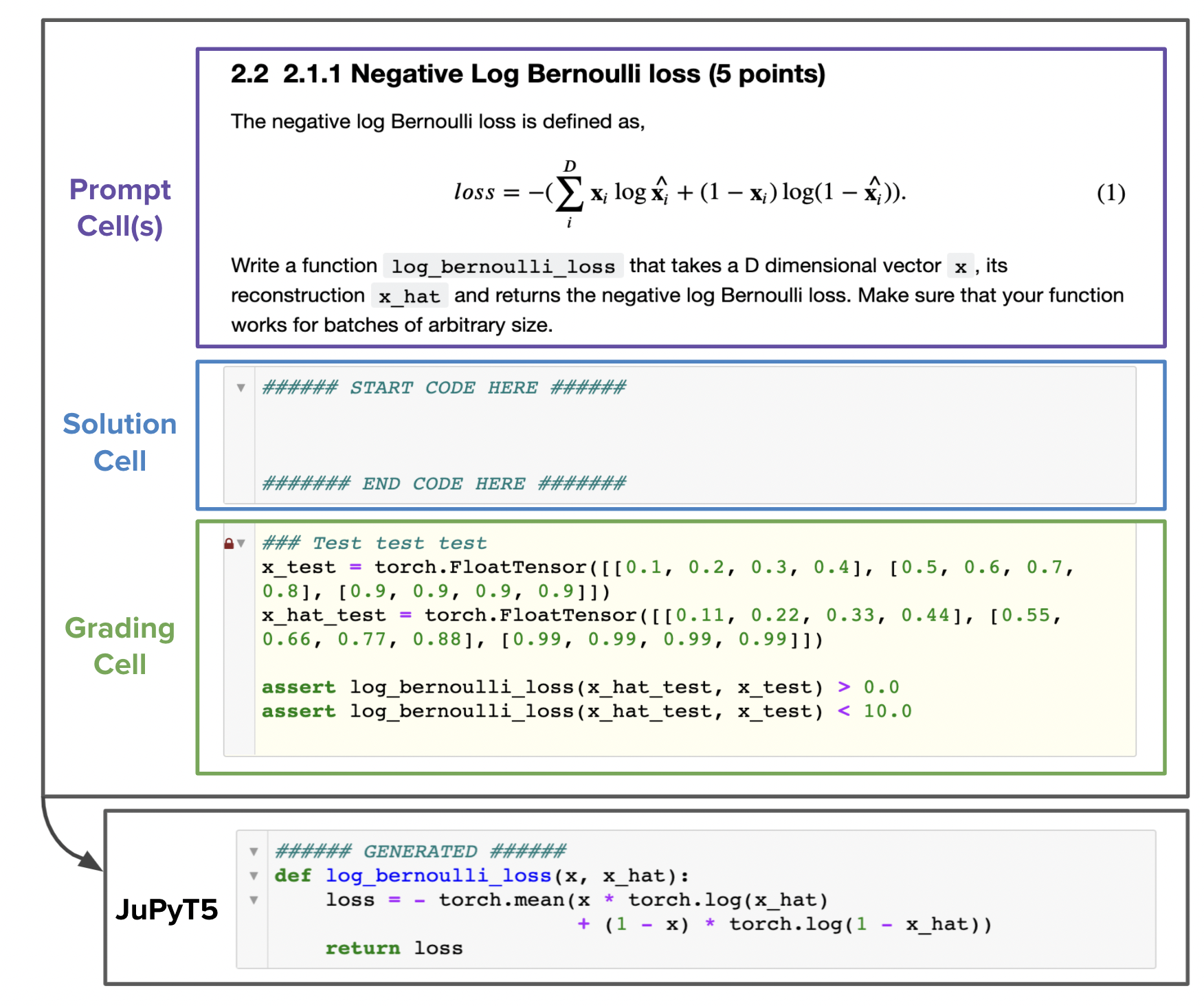}
% \vspace*{-3mm}
\caption{An example (top) from Data Science Problems describing an implementation of a log-Bernoulli loss function using \LaTeX. Our model JuPyT5 correctly interprets the \LaTeX~ definition of the loss as using \texttt{torch} and even names the function and arguments as described in the prompt.}
\label{fig:data-example-2}
%\vspace*{-5mm}
\end{figure}

\section{Datasets}

We train our models on all Jupyter notebooks from all the public GitHub repositories with Jupyter Notebooks as the primary language label as of April 2021, excluding repositories from which our Data Science Problems evaluation were curated. We first discuss the filtering pipeline which yielded 1119 high-quality data science questions with validating unit tests which compose DSP.

\subsection{Data Science Problems}

HumanEval are 164 hand-written programming problems, with accompanying test cases. These involve one-to-one natural language to code pairs, testing reasoning, algorithms, and simple mathematics. The Mostly Basic Programming Problems is a dataset of 974 short Python programs constructed by crowd-sourcing to an internal pool of workers who have basic knowledge of Python. These are single self-contained Python function solving the problem specified. Data Science Problems are 1119 problems from 306 notebooks with an average of 3.6 problems per notebook, and all are executable with data dependencies in a default Anaconda environment. These notebooks are curated from GitHub repositories uploaded by students, which we detect by their usage of the \texttt{nbgrader} notebook grading tool.

\texttt{nbgrader} is a tool used by instructors to create and validate the assignments by executing the code and checking assert statements written by an instructor. Each of these notebooks consists of cells classified as prompt or context cells which define the problem to be solved, solution cells in which the student (or model) should insert an implemented solution, and grading cells which contain unit tests validating the solution. Figure~\ref{fig:data-example} shows one such example: the prompt cell defines the problem (in this case it loads a dataset and suggests modifying zip codes in the Pandas \texttt{dataframe}. The solution cell is to be filled by the student, solving the problem described by the context cell(s) preceding the solution cell. Finally, in most of the cases in \texttt{nbgrader} notebooks, and all of the cases in DSP, the solution cell is followed by a grading cell containing unit tests on the generated code.

To curate the DSP dataset, we start out with a set of 448 Github repositories from the Jupyter Interactive Computing (JuICe)~\cite{Agashe2019JuICeAL} development dataset, which contains of 33K nbgrader notebooks. Using \texttt{nbclient}\footnote{https://github.com/jupyter/nbclient} and a default Anaconda Python 3.9 environment we attempted to execute all of these 33K notebooks. Each cell was limited to 600 seconds of execution and any of the notebooks which violated this limit were discarded. The notebooks which did not include or could not load their data dependencies were also discarded. Any notebook depending on libraries not present in the Python standard library or the Anaconda default data science environment, non-local modules, or any imports which could not be successfully imported were also discarded. 2134 notebooks passed through this filter.

These 2134 notebooks were successfully executable, but did not necessarily implement actual problem grading. Therefore we further filtered the notebooks by checking whether the grading cells following a solution cell had assert statements which referenced the defined method name, function name, variable name or class name from the solution cell. Finally, following this highly stringent criteria to select a notebook and the corresponding solution cell, we identified a subset of 306 notebooks with 1119 solution cells, each of which is preceded by context cells defining the problem and followed by a grading cell testing the functional correctness of the solution. Table~\ref{tab:dsp-stats} shows some high-level statistics of the resulting Data Science Problems dataset.

\begin{table}[htbp]
\caption{Qualitative analysis by understanding 50 examples from Data Science Problems. The various problems range from computing gradient backpropagation, performing complex pandas operations on tables to writing and training scikit-learn models.}
    \centering
    \small
    \begin{tabular}{lll}
        Domain & Examples & Fraction\\
        \midrule
        Math Problems & Compute derivative & 39.5\% \\
        Programming Question & Merge Sort & 26\% \\
        Data Science & Pandas Groupby & 20\% \\
        Machine Learning & Build and train model & 12.5\% \\
        Miscellaneous & HTTP Get Request & 2\% \\
    \end{tabular}
    
    \label{tab:dsp-class-stats}
\end{table}

% \begin{table}[htbp]
%     \centering\footnotesize
%     \begin{tabular}{ll}
%         Import statement & Freq. \\\midrule
%         \texttt{import numpy as np} & 245 \\
%         \texttt{import matplotlib.pyplot as plt} & 143 \\
%         \texttt{import pandas as pd} & 100 \\
%         \texttt{import seaborn as sns} & 49 \\
%         \texttt{from nose.tools import assert\_equals} & 47 \\
%         \texttt{from matplotlib import pyplot as plt} & 43 \\
%         \texttt{import math} & 23 \\
%         \texttt{import ipython} & 22 \\
%         \texttt{import scipy as sp} & 20 \\
%         \texttt{import sklearn as skl} & 18
%     \end{tabular}
%     \caption{Frequency of most common import statements in Data Science Problems.}
%     \label{tab:package-freq}
% \end{table}

\begin{table}[htbp]
    \centering\footnotesize
    \begin{tabular}{ll}
        Modules & Freq. \\\midrule
        \texttt{numpy} & 253 \\
        \texttt{matplotlib} & 180 \\
        \texttt{scipy} & 125 \\
        \texttt{pandas} & 99 \\
        \texttt{sklearn} & 53 \\
        \texttt{seaborn} & 47 \\
        \texttt{nose.tools} & 47 \\
        \texttt{math} & 30 \\
    \end{tabular}
    \caption{Frequency of most common modules used in Data Science Problems.}
    \label{tab:package-freq}
\end{table}

\iffalse
\begin{figure}[htbp]
\centering
\includegraphics[width=\columnwidth]{LaTeX/figs/histogram-packages-dsl.png}
% \vspace*{-3mm}
%\caption{TODO: one unusual suspect nose.assert_equal}
\caption{Frequency of most common packages imported in Data Science Problems.}
\label{fig:histogram-dsl}

%\vspace*{-5mm}
\end{figure}
\fi

\begin{table}[htbp]
    \centering
    \begin{tabular}{ll}
        \midrule
        GitHub Repositories & 69 \\
        Notebooks & 306 \\
        Problem-test pairs & 1119 \\
        Total assert statements & 2298 \\
        Total data files & 92 \\
        Notebooks referencing data & 70 \\
        Problems in data-dependent notebooks & 395\\
        \midrule
    \end{tabular}
    \caption{Statistics of the sources for and features of the Data Science Problems dataset.}
    \label{tab:dsp-stats}
\end{table}

\subsubsection{DSP Problem Analysis}

In order to better understand the contents of the DSP notebooks, we randomly sampled 50 and hand-classified the problems therein. Table~\ref{tab:dsp-class-stats} contains the results of this survey, finding 39.5\% of problems are "math problems" like computing the derivative of a function via back-propagation. 26\% were "programming questions" like implementing merge sort, 20\% were "data science" using Pandas objects and operations on tables, and 12.5\% built, trained and evaluated machine learning models. This survey tracks well with our measurement of 35\% of problems depending on some data file, and 32.5\% of the hand survey of these problems were data science or machine learning. Table~\ref{tab:package-freq} shows the top \texttt{modules} found in DSP notebooks, and they are dominated by plotting packages, math and data science packages like Pandas and SciPy.

\begin{figure}[htbp]
\centering
\includegraphics[width=\columnwidth]{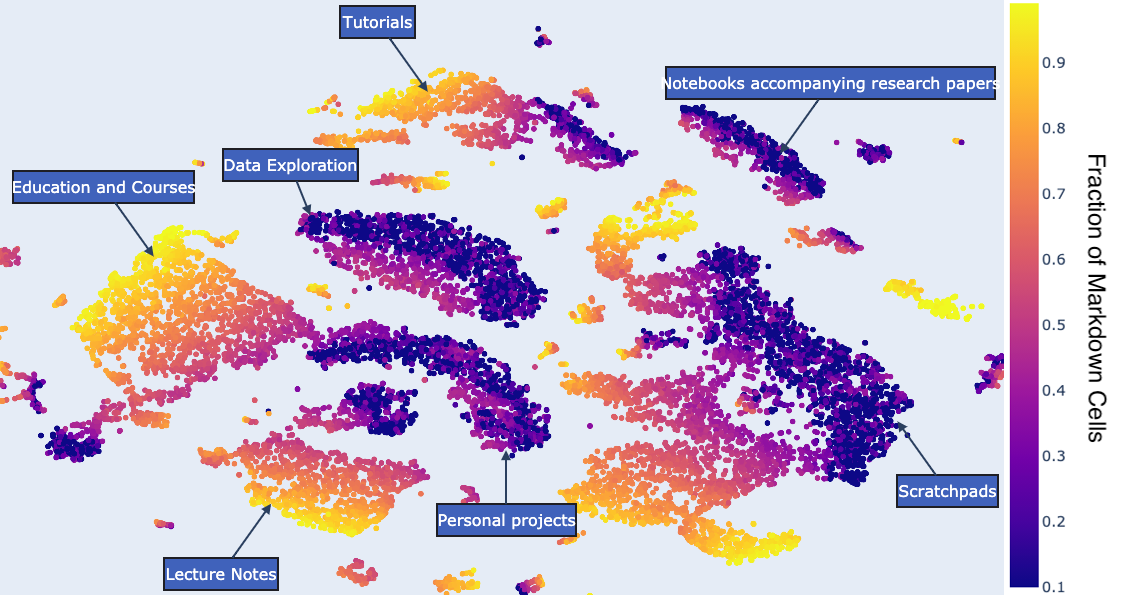}
% \vspace*{-3mm}
\caption{t-SNE Visualization of a sample of 17K notebooks from the pretraining dataset. Cluster labels were assigned by hand after manually observing \~10 members of each cluster.}
\label{fig:tsne}
%\vspace*{-5mm}
\end{figure}

\subsection{Pretraining dataset}

% Add in complete description of processing pipeline and gross statistics of data. Prompt length, cells per notebook, markdown/code fraction cells per notebook, plot markdown fraction for each notebook for example. How many cells had outputs?

Our pre-training and training data consists of all Jupyter Notebooks from all GitHub repositories which were labeled by GitHub as consisting primarily of Jupyter Notebooks, and the repositories were cloned and processed April 2021. In total we obtained 1.97 million repositories and from these derived 9.06 million total Jupyter Notebooks, 7.24 million of which were unique. In order to prevent data leakage into DSP, we removed from our training set all repositories which were in the JuICe testing and development sets, and further ensured no duplicates from these holdout sets were present in our training set.

%In the month of April of 2021, we looked at all the repos on Github with a primary language tag as Jupyter Notebooks and collected all the files in these repositories with a file extension of ".ipynb". In doing so, the data was collected from a total of 1.97M repositories, consisting of a total of 9.06M jupyter notebooks. From this set, we deduplicated all the files which were repeated in our dataset, which amounted to 1.82M notebooks, leaving us with a total of 7.24M notebooks.

%When curating this pretraining dataset, we ensure that none of the 306 notebooks from the DSP dataset leaked into this dataset. With huge pretrained models, one main concern is memorization. Removal of any notebooks on which the downstream DSP evaluation will be performed, ensures in one way, the model do not cheats on the evaluation benchmark we constructed. 

Each notebook consists of a number of cells, with a total number of cells in our corpus being 221 million. Most cells are labeled as a code cell or a Markdown cell by the user; 69.5\% of all the cells, that is 153 million cells are code cells and the rest, 67 million cells, are Markdown cells. Using the whitespace-augmented  byte level byte-pair encoding tokenizer from PyMT5~\cite{clement2020pymt5}, the total number of tokens in the training set is 27.2 billion tokens. Of this, \~38\% is markdown tokens, that is 10.3 billion tokens and the rest of 16.9 billion tokens are code tokens. For reference, Codex was trained on 100 billion total tokens and ~\cite{austin2021program} is trained on 2.81 trillion tokens.

Figure~\ref{fig:tsne}, shows a t-SNE visualization to understand the space of how people use notebooks, so we can judge that our training and evaluation domains are similar. We randomly samples a set of 17K notebook, sampled a subset of cells, trained FastText embeddings, and reduced the dimensionality with t-SNE. Each point in the representation is a single notebook, and the color is determined by the fraction of Markdown cells the notebook possesses. The Markdown content is a clear signal in separating notebooks, as shown in Fig.~\ref{fig:tsne}. Sampling \~10 notebooks from each cluster, we hand-labeled them as shown in the figure. The blue notebooks with low markdown content, have scratchpads, (surprisingly) research code, personal projects. The yellow with high markdown content are pedagogical notebooks, for example tutorials, university assignments.

\subsubsection{Training Subset: Markdown Focused}

Based on the clear separation in the training data between Markdown rich and poor regions, we also elected to train the model on a subset of the notebooks containing `enough' Markdown. Training on this subset essentially is used to test the hypothesis that the model can improve its DSP problem-solving performance by focusing on `literate' code. In subsequent experiments we define the Markdown Focused training subset as notebooks with at least one code cell and at least 1/3 of the cells are Markdown cells. This subset contains 4.1M notebooks, or 3/5 of the total training set, and 15.7B training tokens.

% \subsection{Evaluation dataset}

% {\bf TODO:} Explain "Clean" dataset. Call it something else more descriptive like MD focused or something.

\section{Models}

We use sequence-to-sequence transformers~\cite{vaswani2017attention} of the large BART architecture~\cite{lewis2019bart}, and start all our training with a pre-trained checkpoint from the Python Method Text-to-text Transfer Transformer (PyMT5)~\cite{clement2020pymt5}, using the same training hyperparameters therein.

\subsection{Code Infilling Pretraining}

BART is pre-trained with a span-masking objective, in which spans of tokens are masked in the input, and the objective is to reconstruct these missing spans of tokens in the output. PyMT5 was pre-trained by masking out a syntactically defined part of Python methods (either the signature, docstring, or body) and reconstructing the missing third element. Naturally, as Jupyter notebooks are arranged as code cells, we define the cell-infilling pre-training.

For each cell in each notebook we prepared one source-target example for our sequence-to-sequence model JuPyT5. In our experiments the source was either $C=1$ context cell (we call this the baseline JuPyT5) or $C=3$ context cells directly prior to the target cell, and in our best model case, one extra cell following the target cell (called the cell infilling model). Figure~\ref{fig:data-example} shows what this looks like for $C=1$ previous context cell including the subsequent grading cell in the source.

\subsubsection{Control Codes}

Since the target types in pretraining can be both code and natural language, we add in control codes to indicate to the model which domain it should target, following CTRL~\cite{Keskar2019CTRLAC} and PyMT5. We used fives control tokens, \texttt{<markdown>} and \texttt{<code>} to indicate Markdown and code, respectively, and also added in \texttt{<function>}, \texttt{<class>}, and \texttt{<import>} tokens for other studies not included in this manuscript.

%Text generation have benefited from CTRL tokens~\cite{Keskar2019CTRLAC}, which helps navigate the direction of generation output. In our work, we use five types of CTRL tokens; \<markdown\>, \<function\>, \<class\>, \<import\> and \<code\>. If a cell block is markdown block, it receives a \<markdown\> token, if it contains a function, it received a \<function\> token, and so on. Collisions are resolved by the priority of the list above, example if a code block do not satisfy any criteria, it'll be assigned \<code\> token. 

\subsection{Training Details}

Each JuPyT5 model was trained for 5 epochs (either with the entire training set or the Markdown focused subset) using 80 32GB Tesla V100 GPUs. The hyperparameters for each were kept the same as for PyMT5, except the batch size was changed to accommodate larger batch sizes for data parallelism.

%Language generation in general, and Code generation in specific, is modeled as a left-to-right auto-regressive sequence generation problem. In this work, we rethink this aspect of code generation, deviating away from traditional $$p(x_i|x_{i-1}, x_{i-2}, ..., x_1)$$ conditional probability. Inspired by the success of BERT, where they propose the cloze task of masked language modeling, we propose the Code Infilling objective. 

%Code Infilling extends the BERT objective, from token level mask infilling objective, to a document level code infilling objective. The language modeling objective now becomes, $$p(c_i|c_{1}, ..., c_{i-1}, c_{i+1}, ..., c_t)$$, where $c_i$ represents a code block. In our specific case of Jupyter Notebooks, each $c_i$ represents a cell in the notebook. In essence, the objective of the model becomes, fill in the given code block, given the code blocks above and below. For examples, in Figure X, the model is given the cell prompts, which presents the problem to be solved. Further, it's also presented with the cells in future, specifically in our case, the assert statement against which it'll be graded. Later we'll show, how this information from the future is crucial for the model to learn and solve the problems in DSP. 

%Code Infilling is motivated by observing a simple, yet crucial fact about programming, that code is written in context. In this work, we test out this hypothesis. We model the task as a text-to-text task, where the task is to infilling the masked code blocks, and the goal of the model is to infill the code blocks. 

\section{Experiments and Results}

\subsection{Evaluation Details}

For each problem in DSP, we copied the whole notebook context, replacing only the solution cell of the given problem being solved by JuPyT5. This is because the notebooks have dependencies between cells which can lead to execution failures which are not necessarily the fault of the model if we let its mistake propagate down the notebook. We could define a separate DSP metric in which the model must complete every problem on its own and success or failure of each problem can depend on one another. As we will see, even with this teacher forcing, that is letting the model see the correct solutions to previous problems, the DSP metric remains quite challenging. We leave more permutations of the evaluating the notebooks and their problems to future work.

For DSP a problem is marked passed if and only if the generated code passes the unit test defined in the grading cell below it. We use the pass@$k$~\cite{chen2021evaluating} metric to evaluate the unbiased probability of the model correctly solving the problem in $k$ attempts. For JuPyT5 one attempt is one hypothesis generated by sampling with T=0.8 and nucleus sampling with top-p of 0.95, which was chosen to optimize the HumanEval performance. Note again that the execution environment was a default Anaconda environment with Python 3.9 and all the code in the GitHub repository original hosting the DSP notebook.

\begin{figure}[htbp]
\centering
\includegraphics[width=0.95\columnwidth]{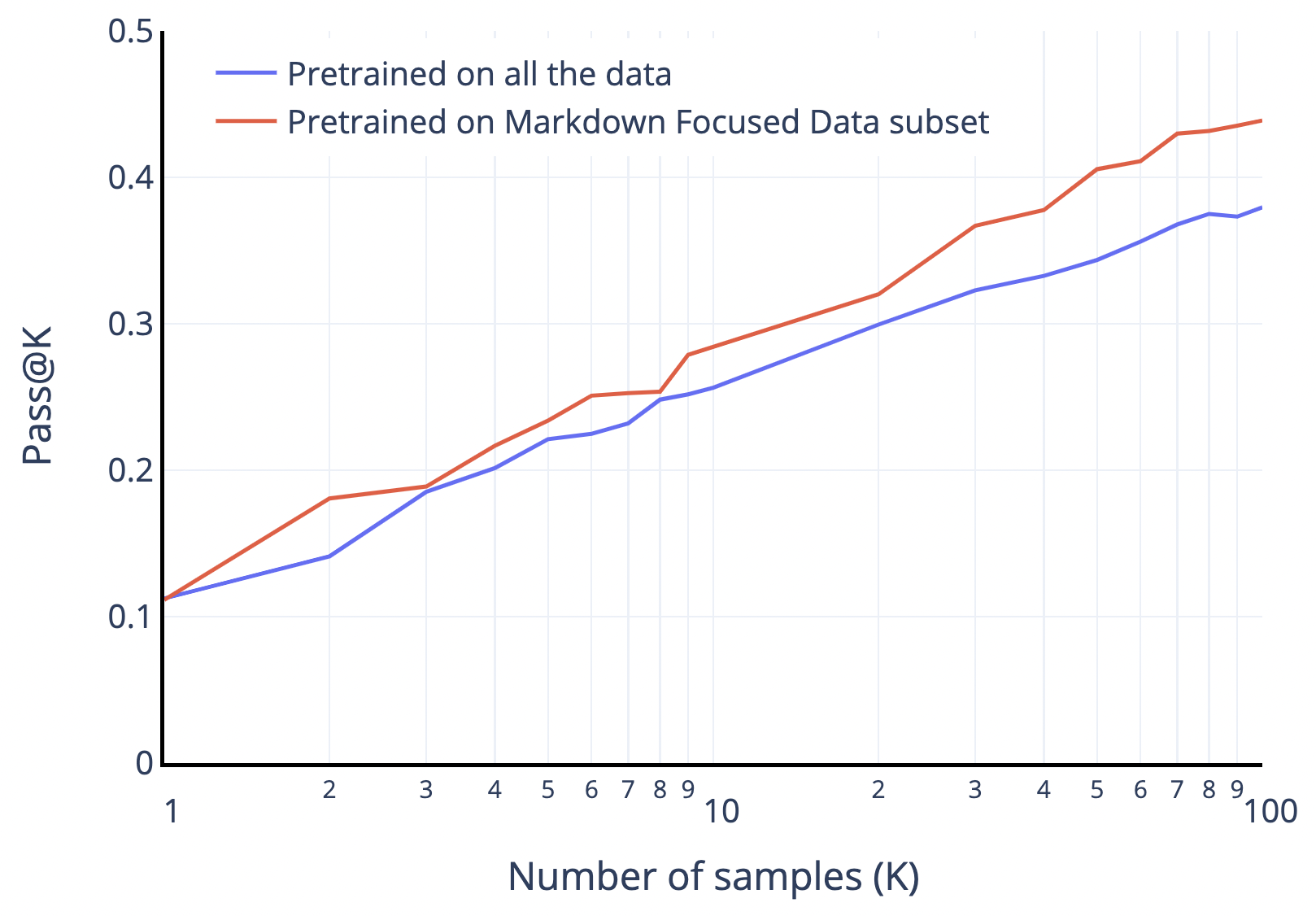}
% \vspace*{-3mm}
\caption{Comparing pass@$k$ rates on DSP for two $C=3$ context models, one trained on the whole dataset and one trained on the Markdown focused subset. We see a modest improvement in pass@$k$ by focusing the training on Markdown-rich data.}
\label{fig:passk-markdown-focused}
%\vspace*{-5mm}
\end{figure}

\begin{figure}[htbp]
\centering
\includegraphics[width=0.95\columnwidth]{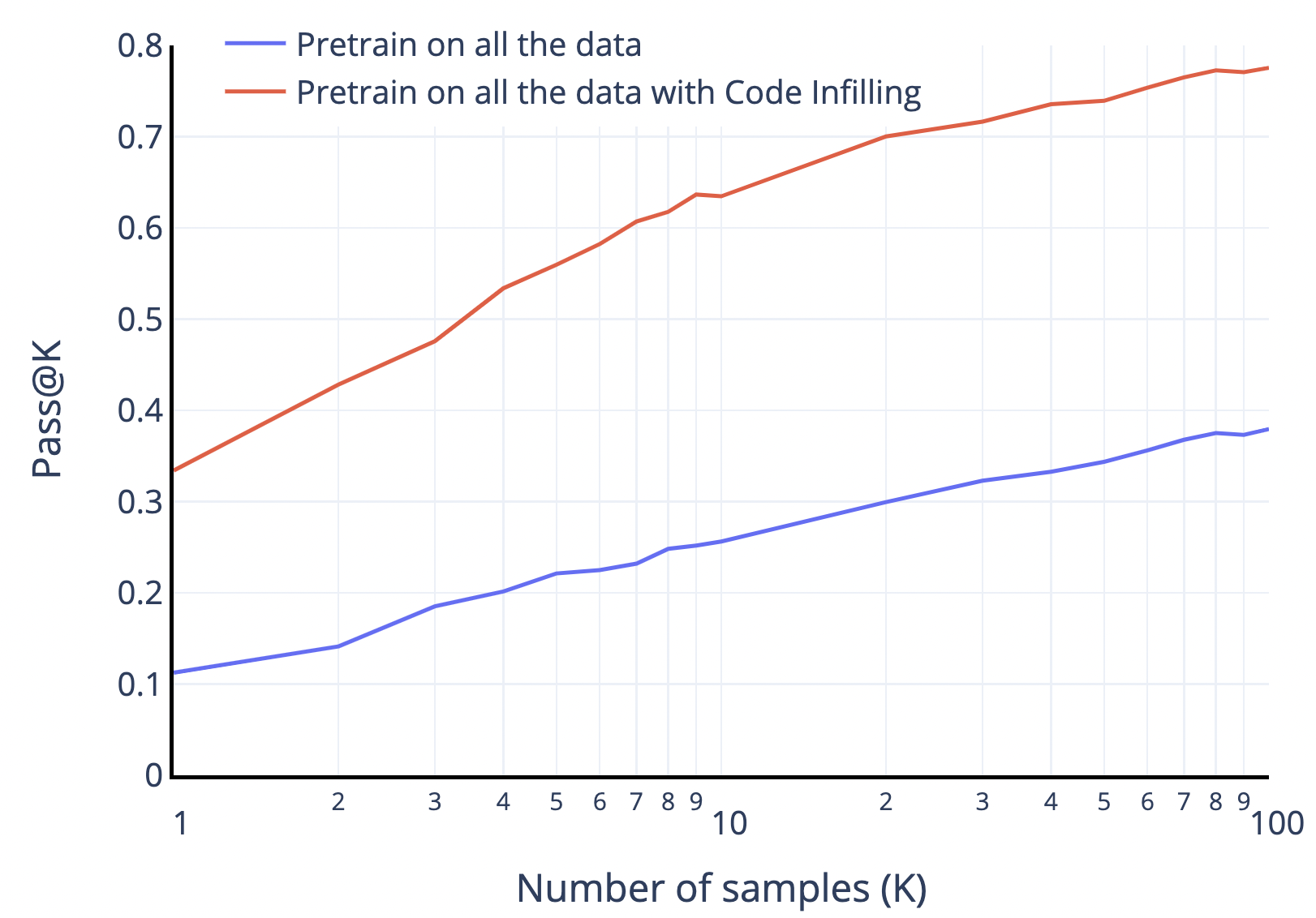}
% \vspace*{-3mm}
\caption{Comparing pass@$k$ rates on DSP for the baseline model and the model trained on code infilling (showing one subsequent cell), showing a dramatic improvement in performance when the model can see the tests.}
\label{fig:passk-infilling}
~\\
\includegraphics[width=0.95\columnwidth]{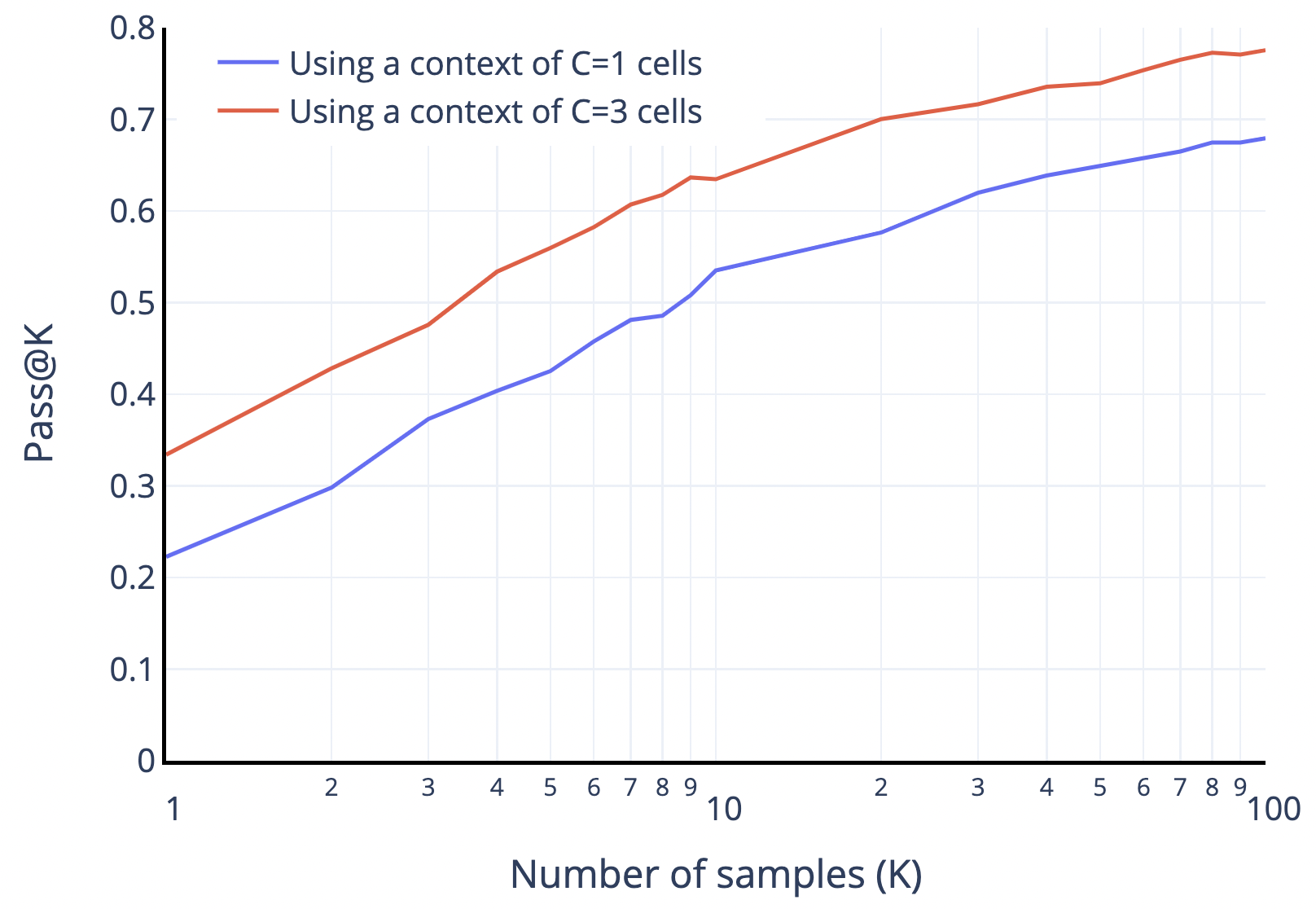}
% \vspace*{-3mm}
\caption{Comparing the pass@$k$ rates on DSP for the cell infilling models (showing the subsequent testing cell) with $C=1$ and $C=3$ previous context cells before the solution cell, showing a consistent \~10\% gain in pass@$k$ for all $k$.}
\label{fig:passk-context}
%\vspace*{-5mm}
%\end{figure}

~\\
%\begin{figure}[htbp]
%\centering
\includegraphics[width=0.95\columnwidth]{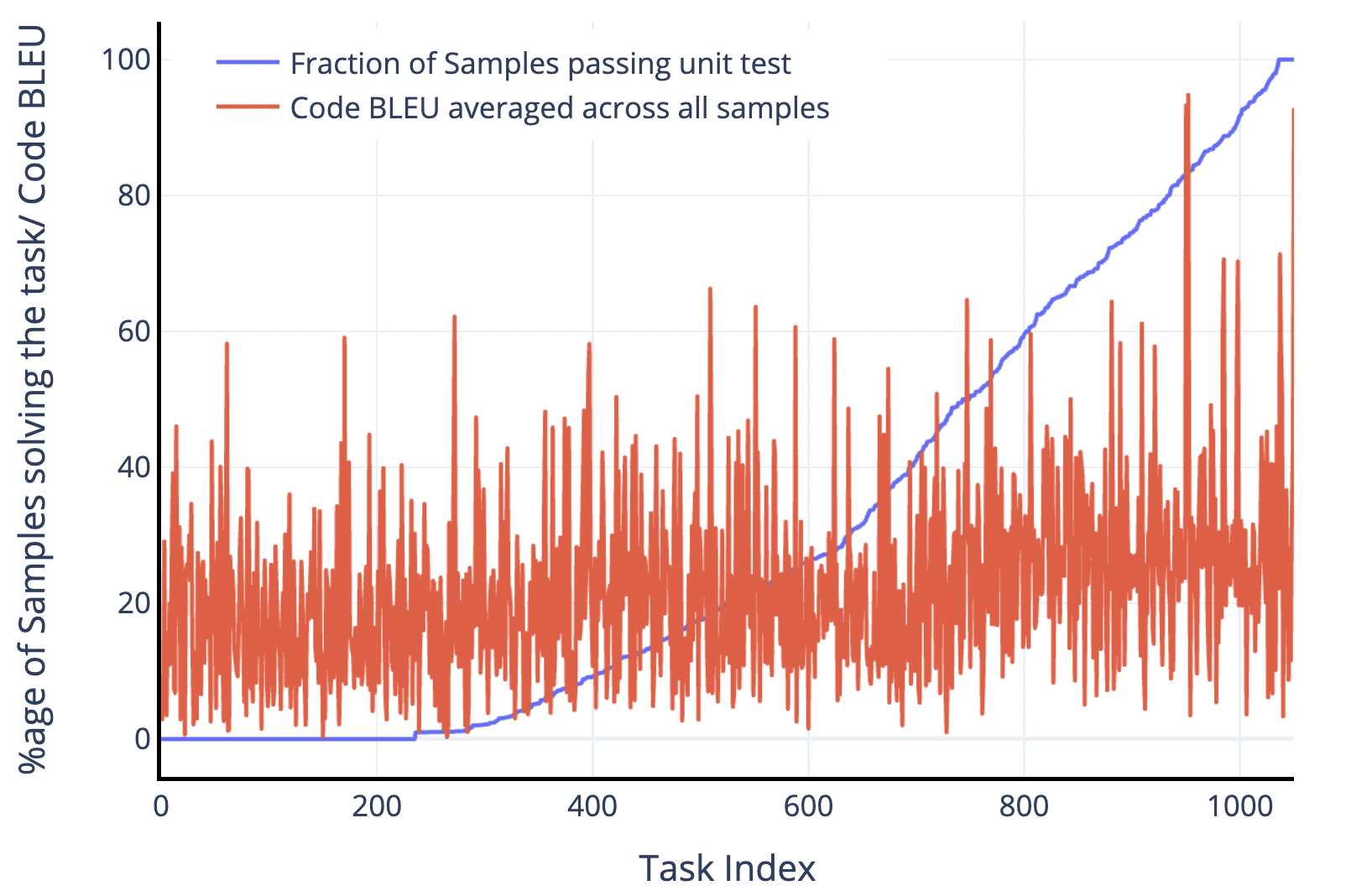}
% \vspace*{-3mm}
\caption{The blue curve is all DSP tasks in order of the JuPyT5 pass rate for 100 samples. The red lines are the average CodeBLEU score between the true code and each hypothesis. There is a weak correlation between the two.
}
\label{fig:bleu-bad}
%\vspace*{-5mm}
\end{figure}

\iffalse
\begin{figure}[htbp]
\centering
\includegraphics[width=\columnwidth]{LaTeX/figs/result-infilling.png}
% \vspace*{-3mm}
\caption{Comparing pass@$k$ rates on DSP for the baseline model and the model trained on code infilling (showing one subsequent cell), showing a dramatic improvement in performance when the model can see the tests its code must pass.}
\label{fig:passk-infilling}
%\vspace*{-5mm}
\end{figure}

\begin{figure}[htbp]
\centering
\includegraphics[width=\columnwidth]{LaTeX/figs/result-context.png}
% \vspace*{-3mm}
\caption{Comparing the pass@$k$ rates on DSP for the cell infilling models (showing the subsequent testing cell) with $C=1$ and $C=3$ previous context cells before the solution cell, showing a consistent \~10\% gain in pass@$k$ for all samples $k$.}
\label{fig:passk-context}
%\vspace*{-5mm}
\end{figure}
\fi

\iffalse % cut figure
\begin{figure}[htbp]
\centering
\includegraphics[width=0.95\columnwidth]{LaTeX/figs/result-select1.png}
% \vspace*{-3mm}
\caption{The red and green curves indicate the scenario when we can generate multiple samples and execute only one. By selecting the sample with highest log-probability per token, we can perform better than selecting a generated sample by chance, but not better than evaluating all hypotheses.}
\label{fig:sample-vs-rankselect}
%\vspace*{-5mm}
\end{figure}
\fi

\subsection{DSP Results}

%\subsubsection{Markdown Focused Pretraining Improves Performance}

Figure~\ref{fig:passk-markdown-focused} shows the result of the first experiment evaluating DSP for JuPyT5 pass@$k$ for $k=1$ to $k=100$. Similar to Codex and \citet{austin2021program} we observe log-linear behavior of the pass rate as a function of the number of attempts. Figure~\ref{fig:passk-markdown-focused} shows two models, one trained on the whole training set, and the other trained on the Markdown focused dataset described above. We see modest gains in performance, most pronounced near $k=100$. As a result of the modest performance improvement with only a 2/5 reduction in data size, we did not dig deeper into this line of inquiry.

\begin{figure}[htbp]
\centering
\includegraphics[width=\columnwidth]{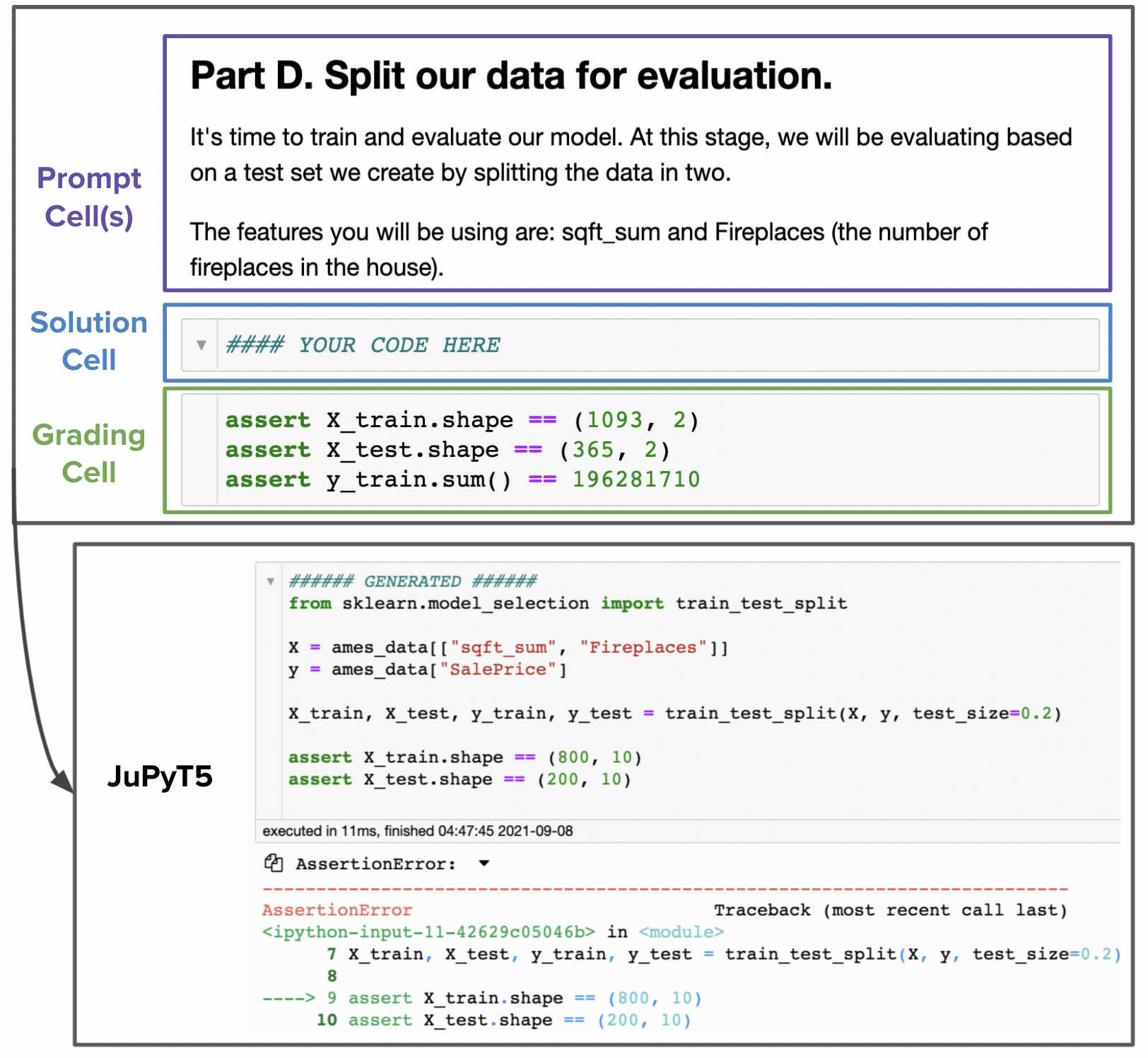}
% \vspace*{-3mm}
\caption{An example of the baseline JuPyT5 predicting assert statements alongside its hypothesis code. This show the prevalence of such unit tests in the training set. One reason the cell infilling (showing the model one subsequent cell) improves performance so much is because the model does not predict further potentially erroneous assert statements when it sees the subsequent grading cell.}
\label{fig:dsp-assert-gen}
%\vspace*{-5mm}
\end{figure}

Figure~\ref{fig:passk-infilling} shows the pass@$k$ rate (with $C=3$) for the baseline model and for the cell infilling model (which sees one additional cell following the target cell). We see a very large improvement in performance, so much that the pass@1 of the cell infilling is comparable to the pass@100 for the baseline. We believe this improvement is for two reasons: the first is in cell infilling the model can see the tests it will be judged by (\citet{austin2021program} found their model performance was much improved by showing the model the tests as well). Our second hypothesis can be best explained in the example of Fig.~\ref{fig:dsp-assert-gen}: when the model does not see subsequent assert statement it often generates them, and they are not always correct. This could be ameliorated by simply ignoring assert statements in the generated hypothesis, something we leave to future work.

Figure~\ref{fig:passk-context} shows our final experiment with two levels of context, $C=1$ and $C=3$ previous context cells before the target. This yields the most consistent boost in performance regardless of the number of samples drawn, which makes sense considering the model can see solutions to some previous problems. We in fact observe this `template modification' behavior in an example generated in Fig.~\ref{fig:dsp-template-example}. The model copies the structure of the code in the prompt cell, even adapting the comments in the function (mostly correctly).

\begin{table}[htbp]
\caption{
We evaluate JuPyT5 on the Data Science Problems with different training schemes and different number of context cells $C$. The baseline is training and evaluating on the previous $C$ cells before the target cell, MD focused is the same sources and targets restricted to the Markdown focused subset. Cell infilling is $C+1$ cells in the source, showing the model one cell past the target cell.
}
\label{tab:dsp-jupyt5}

\footnotesize\centering
\begin{tabular}{llcccc}
\toprule
& & \multicolumn{4}{c}{pass@$k$} \\
& & $1$ & $10$ & $50$ & $100$ \\
\midrule
% Baseline & -\% & -\% & -\% & -\% \\
C=1 & Baseline & 6.5\% &  16.5\% &  22.7\% &  25.3\% \\
 & MD Focused & 7.1\% &  17.3\% &  26.2\% &  27.8\% \\
 & Cell Infilling &  22.3\% &  53.5\% &  65.0\% &  67.9\% \\

\midrule
 C=3 & Baseline & 11.2\% &  25.6\% &  34.4\% &  37.9\% \\
 & MD Focused & 11.2\% &  28.4\% &  40.6\% &  43.9\% \\
 & Cell Infilling &  33.4\% &  63.5\% &  73.9\% &  77.5\% \\

\bottomrule
\end{tabular}
\end{table}

The results of all of these experiments are summarized in Tab.~\ref{tab:dsp-jupyt5} for a few selected $k$ values, and generally reflect our observations above that more context is better, a focused dataset is a modest improvement, and seeing the unit tests is a big boost.

\subsection{HumanEval and MBPP Results}

Table~\ref{tab:humaneval} and Tab.~\ref{tab:mbpp} compare JuPyT5 to baseline models for the Codex HumanEval and MBPP metrics, respectively. We see Codex beats JuPyT5 on HumanEval except when using a much smaller 85M parameter model. This performance gap could be explained by the different formatting between markdown cells and method docstrings (we improved our performance by taking the docstring and presenting it as Markdown). JuPyT5 can beat the Programming Synthesis model at the MBPP metric for all but their largest model. This may not be surprising as the PS model was trained on many English documents which contained some code, and not entirely code like JuPyT5.

\begin{table}[htbp]
\caption{Evaluating JuPyT5 on HumanEval and comparing to Codex.}
\label{tab:humaneval}
\vskip 0.15in
\begin{center}
\begin{small}
%\begin{sc}
\begin{tabular}{lccc}
\toprule
& \multicolumn{3}{c}{pass@$k$} \\
& $k=1$ & $k=10$  & $k=100$ \\
\midrule
Codex-85M & 8.22\% & 12.81\% & 22.4\%  \\
Codex-300M & 13.17\% & 20.37\% & 36.27\% \\
\midrule
JuPyT5-300M & 5.4\% &  15.46\% &  25.6\% \\
\bottomrule
\end{tabular}
%\end{sc}
\end{small}
\end{center}
\vskip -0.1in
\end{table}

\begin{table}[htbp]
\caption{Performance on the MBPP dataset with 80 samples. The fine-tuned Program Synthesis (PS) model is trained on a variety of documents including code documents. We observe that JuPyT5 with 300M parameter count is able to compete with the gigantic 68B PS model.}
\label{tab:mbpp}
\vskip 0.15in
\begin{center}
\begin{small}
%\begin{sc}
\begin{tabular}{lccc}
\toprule
& pass@$k=80$ \\
\midrule
%PS-224M & 7\% \\
PS-422M & 15\% \\
%PS-1B & 21\% \\
PS-4B & 33\% \\
%PS-8B & 40\% \\
PS-68B & 54\% \\
PS-137B & 63\% \\
\midrule
JuPyT5-300M & 52.2\% \\
\bottomrule
\end{tabular}
%\end{sc}
\end{small}
\end{center}
\vskip -0.1in
\end{table}

\subsection{Discussion}

While our best model was able to solve over 77\% of the DSP problems, this is a most optimistic metric as the deployment scenario may not tolerate 100 hypotheses. If users describe their problem and provide test cases however, following the test-driven development model, that could be a scenario in which JuPyT5 is a fairly effective Data Science assistant. The model also seems to effectively bootstrap off of earlier solutions, evidence by the consistent increase in passing performance regardless of samples $k$, and so could become ever more effective as a user develops their program. We did investigate attempting to evaluate only a single hypothesis by choosing the sample with the largest log-likelihood per token, which would support a deployment scenario in which no unit tests are provided, but this offered only a modest average improvement over evaluating a single sample.

\begin{figure}[htbp]
\centering
\includegraphics[width=0.95\columnwidth]{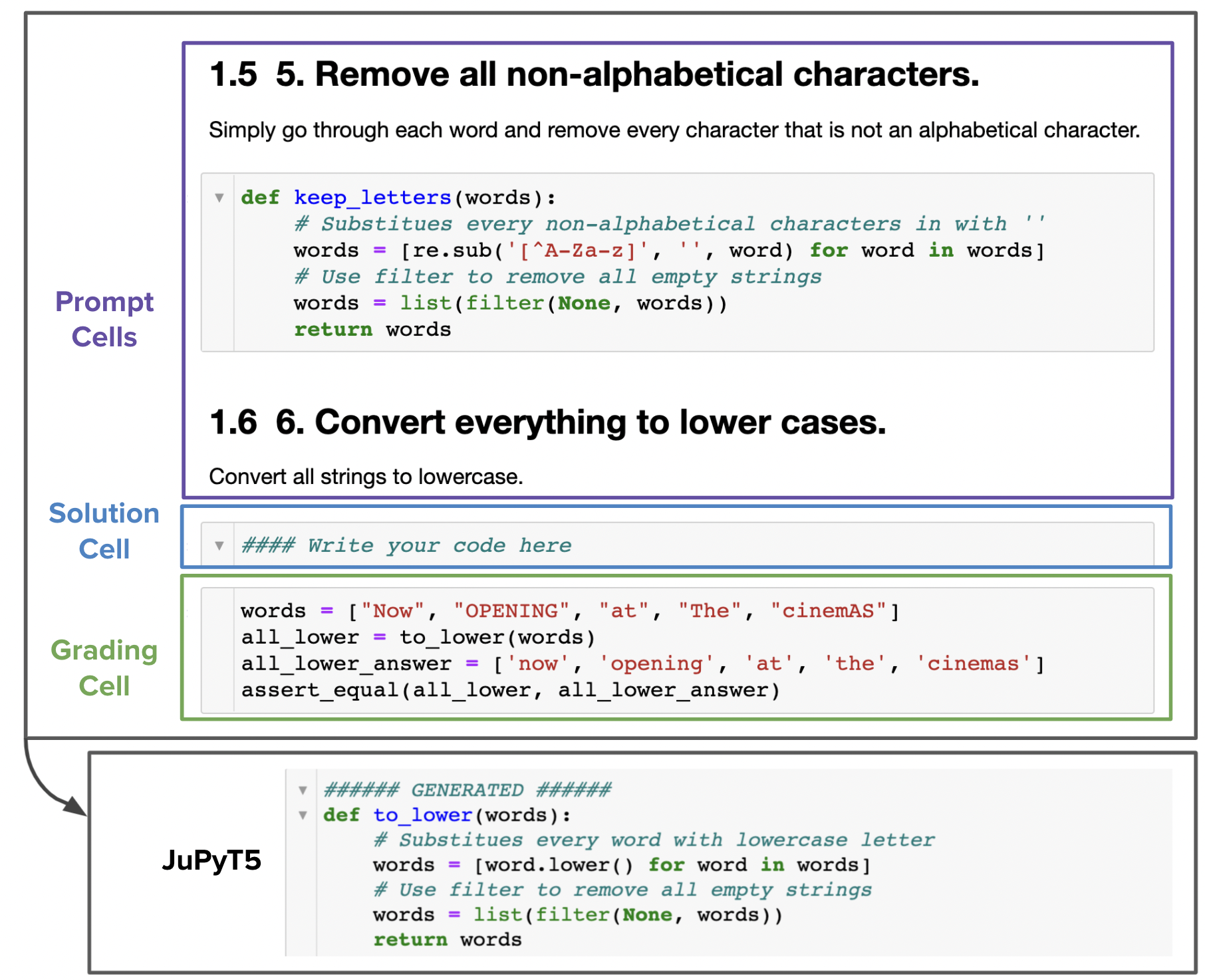}
% \vspace*{-3mm}
\caption{An example of JuPyT5 showing template-like behavior to make its prediction. There is a clear similarity between the prompt code and the solution, the model even mostly correctly adapts the comments.}
\label{fig:dsp-template-example}
%\vspace*{-5mm}
%\end{figure}

%\begin{figure}[htbp]
%\centering
\includegraphics[width=0.95\columnwidth]{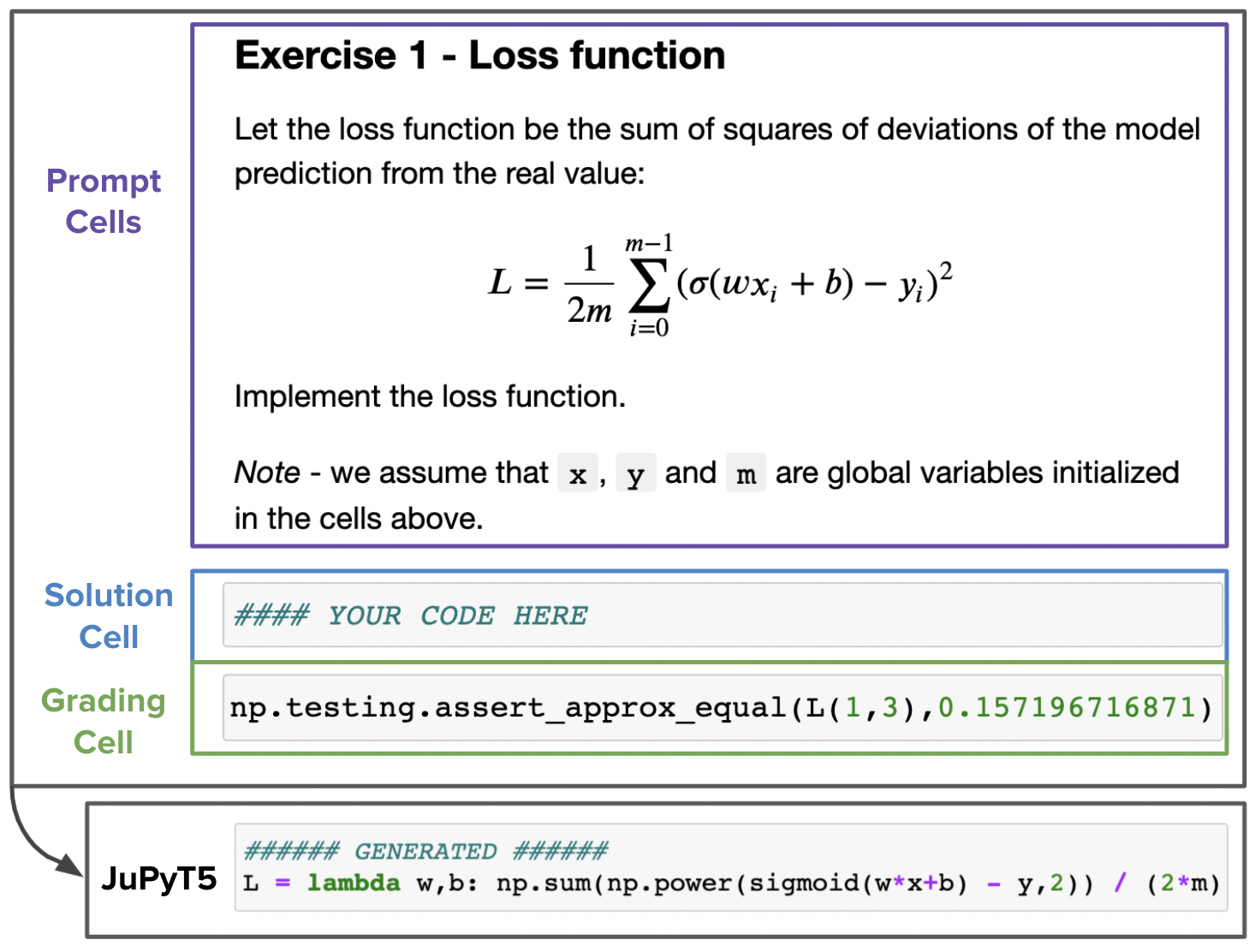} \\
\includegraphics[width=0.95\columnwidth]{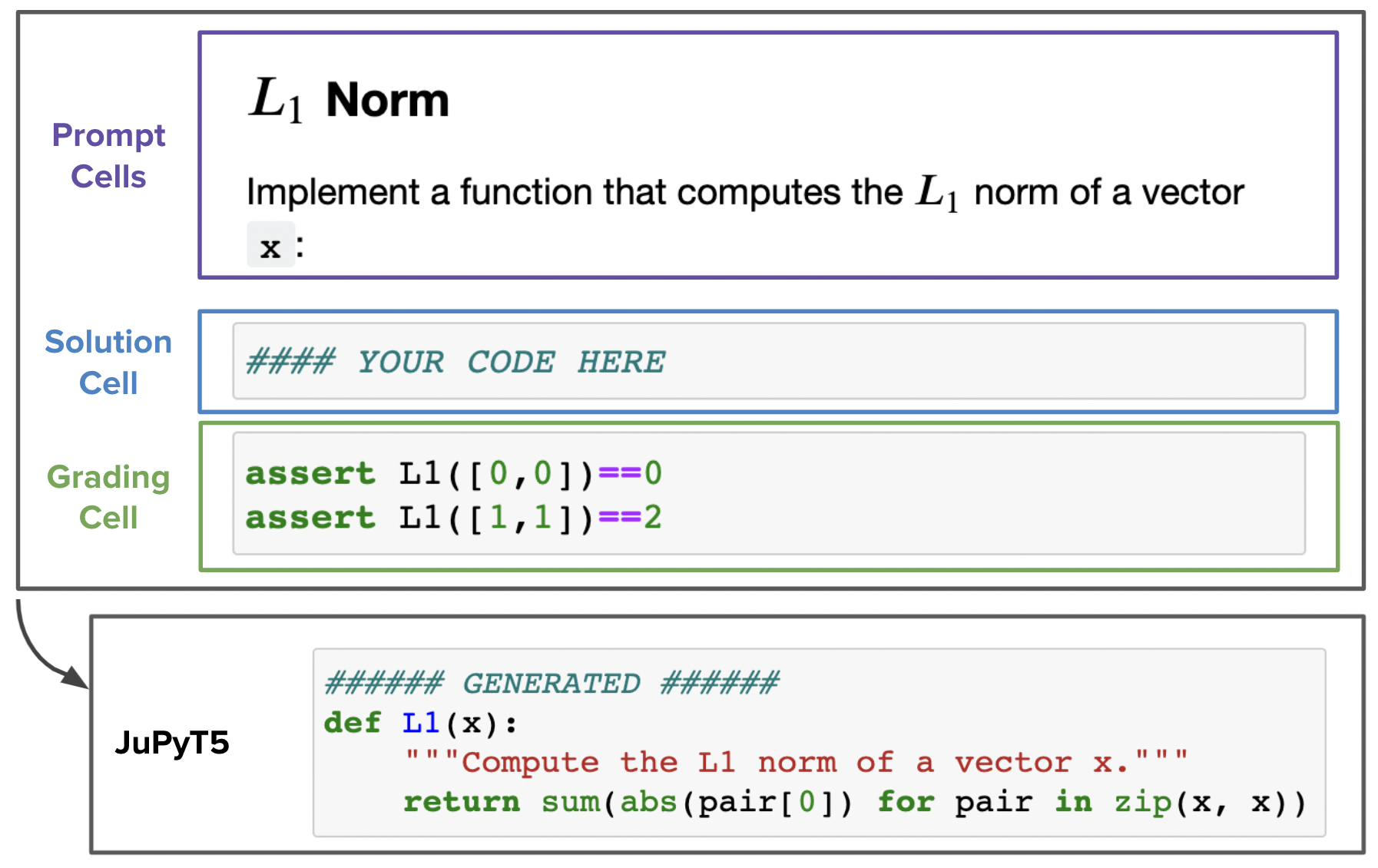}
% \vspace*{-3mm}
\caption{Two `hard' examples from DSP which JuPyT5 rarely solved. The first (top) is a least-squared loss function containing a model, indicating the model struggles composing multiple chains of logic. The second (bottom) is not complicated, but is not defined in the problem, and relies on the model `knowing' the definition of the $L_1$ norm.}
\label{fig:dsp-hard}
%\vspace*{-5mm}
\end{figure}

It can be perhaps most instructive to discuss the `easiest' and `hardest' DSP problems and how the model could solve them. The `easiest' DSP problem consisted mainly of common Pandas \texttt{dataframe} operations like dropping a column. Two of the hardest problems are shown in Fig.~\ref{fig:dsp-hard}. The bottom example problem is to implement the $L_1$ norm, but does not define it like some other DSP example problems, and so the model must rely on having been trained to understand the definition of that norm. This is a scenario which likely can be easily improved by larger model sizes. The top example, however, is at first glance implementing a simple least-squares regression objective, but is posed with a model defined inside. This kind of compound chaining of operations is difficult for these models, a challenge which was also reported by \citet{chen2021evaluating} and \citet{austin2021program}.

Finally we discuss Fig.~\ref{fig:bleu-bad}, which plots the pass rate of each DSP problem in sorted order, along with the average CodeBLEU~\cite{ren2020codebleu} score of the 100 generated samples. We see there is only perhaps a weak correlation between pass rate and CodeBLEU score, showing that BLEU/CodeBLEU are not useful for determining the correctness of hypothesis programs.

\section{Conclusion}

We introduced a new code generation evaluation metric called Data Science Problems consisting of over 1000 problems, many of which depend on data dependencies, and all of which are executable with unit tests. We train a new model, JuPyT5 on almost all publicly available Jupyter Notebooks, and show it is capable of solving over 77\% of the problems. While this is an optimistic estimate, we believe this proves the feasibility of a data science assistant in the form of these large transformer models. While it is clear from the literature that larger models can solve more problems, challenges in complex code synthesis remain, and the DSP benchmark can help our community of researchers to overcome these modeling challenges.

\bibliography{aaai22,emnlp2020,ewashreferences}

\end{document}